\def\year{2020}\relax
\documentclass[letterpaper]{article}
\usepackage{aaai20}
\usepackage{times}
\usepackage{helvet}
\usepackage{courier}
\usepackage{url}
\usepackage{graphicx}
\usepackage{arydshln}
\usepackage{amssymb}
\usepackage[linesnumbered,ruled]{algorithm2e}
\usepackage{multirow}     
\usepackage{multicol} 
\usepackage{booktabs}
\usepackage{mathtools}

\usepackage{subcaption}

\usepackage{pdfpages}
\nocopyright

\newcommand{\citet}[1]
{\citeauthor{#1}~\shortcite{#1}}
\newcommand{\citep}{\cite}

\DeclareMathOperator*{\EE}{\mathbb{E}}

\title{A Causal Inference Method for Reducing Gender Bias\\in Word Embedding Relations}
\author{
Zekun Yang \and Juan Feng\thanks{Corresponding author.}\\
Department of Information Systems, College of Business\\City University of Hong Kong\\Hong Kong SAR, China\\
zekunyang3-c@my.cityu.edu.hk, juafeng@cityu.edu.hk
}

\begin{document}
\maketitle

\begin{abstract}
Word embedding has become essential for natural language processing as it boosts empirical performances of various tasks. However, recent research discovers that gender bias is incorporated in neural word embeddings, and downstream tasks that rely on these biased word vectors also produce gender-biased results. While some word-embedding gender-debiasing methods have been developed, these methods mainly focus on reducing gender bias associated with gender direction and fail to reduce the gender bias presented in word embedding relations. In this paper, we design a \emph{causal} and \emph{simple} approach for mitigating gender bias in word vector relation by utilizing the statistical dependency between gender-definition word embeddings and gender-biased word embeddings. Our method attains state-of-the-art results on gender-debiasing tasks, lexical- and sentence-level evaluation tasks, and downstream coreference resolution tasks.
\end{abstract}

\section{Introduction}

Word embeddings that capture the relationship between words have become an essential component of natural language processing (NLP) due to its enhancement to the performance of numerous downstream tasks including sentiment analysis \citep{tang2014learning}, information retrieval \citep{ganguly2015word}, and question answering \citep{yang2016stacked}.

Recent studies have discovered that neural word embeddings, such as Word2Vec \citep{Mikolov2013} and GloVe \citep{Pennington2014}, exhibit gender bias to various extents \citep{bolukbasi2016man,zhao2018learning}. Sources of gender bias of word vectors are two-fold. Firstly, word vectors of gender-neutral words are found to favor spatial directions of either male- or female-definition words, which is defined as \emph{gender bias associated with gender direction}. Secondly, gender bias is also presented in the distance between embeddings of gender-neutral words, which is attributed to as \emph{gender bias in word vector relation}.

Gender-biased word embeddings lead to serious social implications. For example, many NLP pipelines routinely use clustering algorithms as a feature extraction step. Yet, gender-biased word embeddings are known to be susceptible to clustering algorithms. For instance, \citet{gonen2019lipstick} discover that clustering algorithms tend to differentiate word vectors as either male or female associated, despite the fact that these words are by definition not associated with gender. When it comes to concrete downstream NLP tasks, the usage of gender-biased word vectors is even more worrying. \citet{zhao2018gender} illustrate that the word embedding component used in coreference resolution systems results in gender-biased identification outcomes. \citet{park2018reducing} find that one of the leading causes of gender bias in abusive language detection models is the sexist word embeddings employed.

Due to the importance of eliminating gender bias from word embeddings, previous research has proposed post-processing algorithms \citep{bolukbasi2016man,Kaneko:ACL:2019} and word-vector-learning algorithms \citep{zhao2018learning} aiming at solving this issue. While these methods adopt the strategy of forcing gender-biased word vectors to be orthogonal to a gender directional vector, which alleviates the gender bias associated with gender direction significantly, this strategy has been proven to be unhelpful in reducing the gender bias in word vector relation \citep{gonen2019lipstick}. Moreover, some previous methods require tuning several hyper-parameters as well as differentiating between gender-neutral and gender-biased words.

In this paper, we design a \emph{causal} and \emph{simple} post-processing approach for reducing gender bias in word embeddings, which addresses the limitations of previous research. While traditional methods mitigate gender bias by diminishing the relationship between the gender-biased word vector and the gender direction, our method is quite different: we propose to learn the spurious gender information via \emph{causal inference} by utilizing the statistical dependency between gender-biased word vectors and gender-definition word vectors. The learned spurious gender information is then subtracted from the gender-biased word vectors to achieve gender-debiasing. Through learning the gender information in gender-biased word vectors, \emph{both} gender bias in word vector relation and gender bias associated with gender direction are approximated and subtracted. The proposed method is \emph{theoretically interpretable} via the Half-Sibling framework proposed by \citet{Scholkopf2016}, and it is \emph{practically simple} as only one hyper-parameter needs to be tuned, and no additional classification of words is required.

The experimental results on gender direction relation tasks, gender-biased word relation tasks, lexical- and sentence-level evaluation tasks, and downstream coreference resolution tasks show the effectiveness of our proposed method. Specifically, while constantly achieving the best performance on the gender-biased word relation tasks GBWR-Correlation and GBWR-Profession (see Section 4.2), our proposed method attains 6.72\% and 4.98\% improvement respectively compared to the runner-up method, and it results in 16.89\% and 17.02\% improvement compared to the original embedding. Furthermore, our method achieves an average improvement of 6.96\% on eight lexical-level evaluation tasks, and it obtains an average improvement of 14.19\% on 20 sentence-level evaluation tasks compared to the prior best post-processing methods. Lastly, on the downstream gender coreference resolution task WinoBias, our method reduces the difference between the coreference accuracy of pro-stereotyped and anti-stereotyped sentences by 87.26\% compared to the runner-up method.

The contributions of this paper are summarized as follows: (1) To the best of our knowledge, our method is the \emph{first} word-vector gender-debiasing method that applies \emph{causal inference} to learn and subtract spurious gender information; (2) The method we propose is light-weight and simple to implement; (3) The experimental results demonstrate that our method is the \emph{state-of-the-art} post-processing method that mitigates \emph{both} types of gender biases, enhances the general quality of word vectors, and diminishes gender bias in downstream tasks.
 
The rest of the paper is organized as follows. First, we briefly review related work on gender bias identification and gender-debiasing methods for word vectors. Then, we introduce our Half-Sibling Regression method for alleviating gender bias in word embeddings. Finally, we conduct experiments on a series of gender-debiasing tasks, lexical- and sentence-level evaluation tasks, and coreference resolution tasks to reveal the effectiveness of the proposed post-processing algorithm\footnote{Codes are available at \url{https://github.com/KunkunYang/GenderBiasHSR}}.

\section{Related Work}

\subsection{Aspects of Gender Bias}

Words can be classified as gender-definition and non-gender-definition words. Gender-definition words are associated with gender by definition, such as $mother$ and $father$. While non-gender-definition words are not associated with gender, the word embeddings of some non-gender-definition words are not gender-neutral and contain gender bias. Previous research has discovered and defined two types of gender biases in word vectors: gender bias associated with gender direction and gender bias in word vector relation.

\paragraph{Gender bias associated with gender direction} Gender bias associated with gender direction is the bias with respect to a vector that mostly captures the gender information in a word embedding. The gender direction is defined as the difference between the word vectors of male-definition words (e.g. $he$) and female-definition words (e.g. $she$), in which the vector $\overrightarrow{he}-\overrightarrow{she}$ is the most widely accepted definition of the gender direction. \citet{bolukbasi2016man} discover that some words have a large projection on the gender direction, and this large projection value is considered as the gender bias associated with gender direction. For example, given the word embedding GloVe pre-trained on 2017 January dump of English Wikipedia, the cosine similarity between $\overrightarrow{nurse}$ and the gender direction $\overrightarrow{he}-\overrightarrow{she}$ has a negative value of -0.2146, and the cosine similarity between $\overrightarrow{colonel}$ and the gender direction has a positive value of 0.1830. For reference, $\overrightarrow{tree}$, which is seldom related to human, has a cosine similarity of 0.0046 with respect to the gender direction, which is much closer to zero compared to $\overrightarrow{nurse}$ and $\overrightarrow{colonel}$. Therefore, it is easy to see that $\overrightarrow{nurse}$ has a female bias, and $\overrightarrow{colonel}$ has a male bias.

The gender bias associated with gender direction of a word embedding is removed when the projection of the word embedding on the gender direction is zero.

\paragraph{Gender bias in word vector relation} While much previous research believes that the gender bias associated with gender direction represents all gender-stereotypical information in word embeddings, \citet{gonen2019lipstick} show that after diminishing the cosine similarity between the word embedding and the gender direction to zero, another systematic gender bias still exists, which is the gender bias implied in the relations between the non-gender-definition word vectors (referred to as \emph{gender bias in word vector relation}). After subtracting the gender direction, most gender-biased embeddings of non-gender-definition words remain their previous similarity to each other, and they still cluster together as male- and female-biased words. For example, in the pre-trained GloVe embedding, the cosine similarity between two female-biased word vectors $\overrightarrow{dancer}$ and $\overrightarrow{nurse}$ is 0.2728, while the cosine similarity between $\overrightarrow{dancer}$ and the male-biased word $\overrightarrow{colonel}$ has a much smaller value of 0.0891. This close relationship between the two female-biased word vectors $\overrightarrow{dancer}$ and $\overrightarrow{nurse}$ cannot be thoroughly eliminated by just removing their relation to the gender direction.

To eliminate gender bias presented in word vector relations, the similarity between male/female-biased word embeddings needs to be removed. \citet{gonen2019lipstick} propose five tasks to measure the degree of gender bias in word vector relation.

\subsection{Prior Gender-Debiasing Methods}

After defining what gender bias is, we now look at prior gender-debiasing methods. Word-embedding gender-debiasing methods can be classified into two groups: post-processing approaches and word-vector-learning approaches.

Post-processing approaches alleviate gender bias during the post-processing period of word vectors. \citet{bolukbasi2016man} develop a hard-debiasing method which first classifies the words into gender-definition words and non-gender-definition words using a support vector machine classifier trained on a seed set of gender-definition words, and then projects non-gender-definition words to a subspace that is orthogonal to a gender directional vector defined by a set of male- and female-definition word pairs. \citet{Kaneko:ACL:2019} construct a gender-preserving debiasing method that utilizes an autoencoder. The loss function of the autoencoder retains the gender-related information of gender-definition words, requires the gender-biased and gender-neutral words to be embedded into a subspace which is orthogonal to a gender directional vector computed in a similar fashion as in \citet{bolukbasi2016man}, and minimizes the reconstruction loss of the autoencoder.

Furthermore, debiasing methods that mitigate gender bias during the learning process of word vectors have also been developed. \citet{zhao2018learning} propose Gender-Neutral GloVe, a word-vector-learning algorithm in which the objective function is a combination of the original GloVe objective function and two extra components. One of the extra components maximizes the distance between male- and female-definition words, the other component requires the non-gender-definition words to be orthogonal to a gender directional vector defined similarly as in \citet{bolukbasi2016man}.

While both post-processing approaches and word-vector-learning approaches could effectively mitigate gender bias, some limitations exist. Firstly, all three previous methods reduce gender bias by forcing the non-gender-definition word vectors to be orthogonal to a gender directional vector. This strategy mitigates the gender bias associated with gender direction efficiently by directly removing spurious associations between gender-biased word vectors and the gender direction. Yet, according to the experimental results of \citet{gonen2019lipstick}, this strategy fails to significantly reduce the gender bias in word vector relation. Secondly, for Gender-Neutral GloVe and the gender-preserving debiasing method, as two to three constraints are added to the original objective function, the hyper-parameters of these constraints need to be optimized, which complicates the debiasing process. Thirdly, for the gender-preserving debiasing method, the distinction between gender-neutral words and gender-biased words in non-gender-definition words is necessary for the undersampling process, resulting in an extra classification step compared to the other methods.

\section{Causal Gender-Debiasing}

To comprehensively reduce gender bias, we propose a post-processing algorithm that builds upon a statistical dependency graph and uses causal inference method to \emph{learn} and \emph{subtract} the spurious gender information contained in non-gender-definition word vectors directly, instead of \emph{only} removing the relationship between the gender direction and the word vectors.

\begin{figure}[ht]
\centering
\includegraphics[width = 0.45\textwidth]{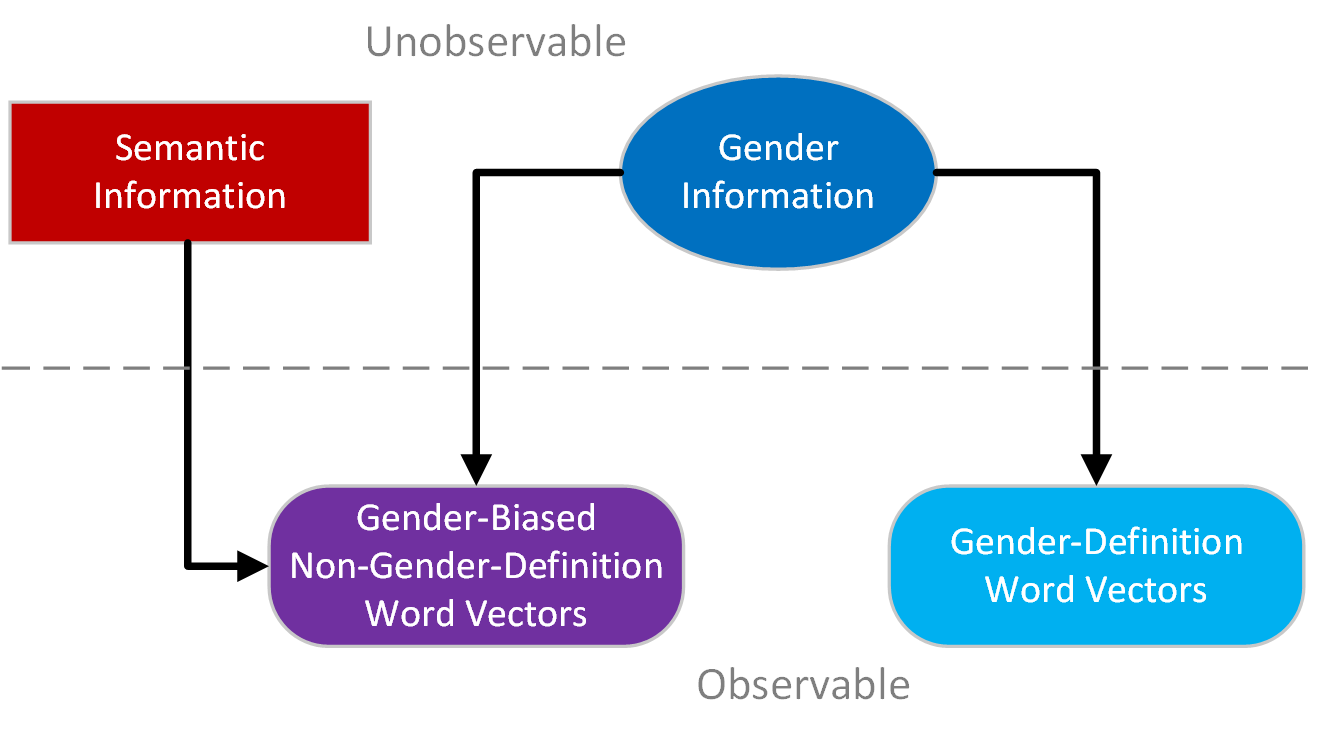}
\caption{Relation between gender-definition word vectors and gender-biased non-gender-definition word vectors}
\label{fig:causalGraph}
\end{figure}

Consider two classes of words, gender-definition word vectors that embody gender information by definition, and non-gender-definition word vectors that contain gender bias, like $\overrightarrow{nurse}$ and $\overrightarrow{colonel}$. It is easy to see that both word vectors contain underlying gender information. In addition, the semantic content (apart from gender information) of the two groups of word vectors is different. While non-gender-definition word vectors contain abundant semantic content, the gender-definition word vectors contain little semantic content apart from the gender information. Figure \ref{fig:causalGraph} illustrates the relation between the above-mentioned groups of word vectors and the underlying contents, in which the black arrows indicate statistical dependency between variables.

Based on the semantic features of gender-definition word vectors and gender-biased non-gender-definition word vectors, we could formulate a procedure to extract the desired gender-debiased semantic information of non-gender-definition word vectors. Specifically, we propose that the debiased non-gender-definition word vectors $\hat{\mathbf{V}}_N$ is learned by subtracting the approximated gender information $\hat{\mathbf{G}}$ from the original non-gender-definition word vectors $\mathbf{V}_N$:
\begin{equation} \label{eq:one}
\hat{\mathbf{V}}_N \coloneqq \mathbf{V}_N - \hat{\mathbf{G}},
\end{equation}
where the approximated gender information $\hat{\mathbf{G}}$ is obtained by predicting $\mathbf{V}_N$ using the gender-definition word vectors $\mathbf{V}_D$:
\begin{equation} \label{eq:two}
\hat{\mathbf{G}} \coloneqq \EE [\mathbf{V}_N | \mathbf{V}_D].
\end{equation}
Since $\mathbf{V}_N$ and $\mathbf{V}_D$ embody the same gender information, when predicting $\mathbf{V}_N$ using $\mathbf{V}_D$, the underlying gender information is learned by $\hat{\mathbf{G}}$. Furthermore, as $\mathbf{V}_D$ contains little semantic information apart from the gender information, when approximating $\mathbf{V}_N$ using $\mathbf{V}_D$, the semantic information of $\mathbf{V}_N$ is not learned by $\hat{\mathbf{G}}$. Hence, when we subtract $\hat{\mathbf{G}}$ from $\mathbf{V}_N$, only spurious gender information is eliminated, and the semantic information of $\mathbf{V}_N$ is preserved, which is eventually the gender-debiased word embeddings.

\begin{algorithm}[ht]
\SetKwInOut{Input}{Input}
\SetKwInOut{Output}{Output}
\Input{ Matrix $\mathbf{V}_D$ of gender-definition word vectors as columns, Matrix $\mathbf{V}_N$ of non-gender-definition word vectors as columns, Ridge Regression constant $\alpha$.} 

Compute the weight matrix of Ridge Regression: $\mathbf{W} \leftarrow \left  ((\mathbf{V}_D)^\top \mathbf{V}_D  + \alpha  \mathbf{I} \right )^{-1} (\mathbf{V}_D)^\top \mathbf{V}_N$

Compute the approximated gender information: $\hat{\mathbf{G}} \leftarrow \mathbf{V}_D \mathbf{W}$

Subtract gender information from the non-gender-definition word vectors: $\hat{\mathbf{V}}_N \leftarrow \mathbf{V}_N - \hat{\mathbf{G}}$

\Output{HSR debiased non-gender-definition word vectors $\hat{\mathbf{V}}_N$.}

\caption{HSR for gender-debiasing}
\label{alg:hsr}
\end{algorithm}

The above debiasing framework is named Half-Sibling Regression (HSR) as we inherit the idea from a confounding-noise-elimination method proposed by \citet{Scholkopf2016}. The name of the framework comes from the fact that $\mathbf{V}_N$ and $\mathbf{V}_D$ share one parent of gender information, but their other parent, semantic information, is different. Therefore, $\mathbf{V}_N$ and $\mathbf{V}_D$ are half-siblings. HSR has proven to be theoretically effective. \citet{Scholkopf2016} show that HSR approximates the cleaned variable not less than the mean-subtraction method does.

To calculate $\hat{\mathbf{G}}$, we use Ridge Regression \citep{hoerl1970ridge}. The formal procedure of our post-processing algorithm is described in Algorithm \ref{alg:hsr}. Through learning the approximated gender information $\hat{\mathbf{G}} $ in gender-biased word embeddings, both gender bias in word vector relation and gender bias associated with gender direction are learned and subtracted. Our proposed algorithm is not only \emph{theoretically interpretable} but also \emph{practically simple}. HSR achieves gender-debiasing by utilizing the causal assumptions between word embeddings (see Figure \ref{fig:causalGraph}), which ensures the interpretability of the model. Moreover, HSR has only one hyper-parameter (Ridge Regression constant) that needs to be tuned, and it does not require extra classification.

When implementing Algorithm \ref{alg:hsr}, for the gender-definition words, we combine the male- and female-word lists\footnote{\url{https://github.com/uclanlp/gn_glove}} (223 words each) in \citet{zhao2018learning} as the gender-definition word list. Words outside of the gender-definition word list are considered as non-gender-definition words. Furthermore, for the Ridge Regression constant $\alpha$ in Half-Sibling Regression, we fix $\alpha = 60$ throughout the experiment as it works well in practice.

\section{Experiments}

To show that our proposed HSR post-processing method reduces \emph{both} gender bias associated with gender direction and gender bias in word vector relation, we test our method on \emph{both} gender direction relation tasks and gender-biased word relation tasks. Furthermore, we show that the quality of the word embedding is not only preserved but also enhanced after adopting our method by testing it on lexical- and sentence-level evaluation tasks. Lastly, we show that the word vectors generated by our method reduces gender bias in the downstream task coreference resolution.

We examine our proposed HSR method on GloVe \citep{Pennington2014} pre-trained on 2017 January dump of English Wikipedia (represented by HSR-GloVe), and compare the performance of our algorithm with previous post-processing methods including the hard-debiasing method (Hard-GloVe) \citep{bolukbasi2016man} and the gender-preserving debiasing method (GP-GloVe) \citep{Kaneko:ACL:2019}. Additionally, we also compare with the word-vector-learning (shortened to WV-Learning in the tables) method for gender-debiasing: Gender-Neutral GloVe (GN-GloVe) \citep{zhao2018learning}. We use pre-trained original GloVe embedding\footnote{\url{https://github.com/uclanlp/gn_glove}} as well as the pre-trained GP-GloVe\footnote{\url{https://github.com/kanekomasahiro/gp_debias}} and GN-GloVe\footnote{\url{https://github.com/uclanlp/gn_glove}} embedding released by the original authors.

\subsection{Gender Direction Relation}

Gender direction relation tasks aim to measure the degree of gender bias associated with gender direction (see Section 2.1 for more details).

\begin{table}[ht]
  \centering
  \caption{Result of gender direction relation tasks}
\scalebox{0.62}{%
    \begin{tabular}{lrrrrr}
     \toprule
          &       & \multicolumn{3}{c}{\textbf{Post-Processing}} & \multicolumn{1}{c}{\textbf{WV-Learning}} \\
           \cmidrule(r){2-5} \cmidrule(r){6-6}
          & \multicolumn{1}{r}{\textbf{GloVe}} & \multicolumn{1}{r}{\textbf{Hard-GloVe}} & \multicolumn{1}{r}{\textbf{GP-GloVe}} & \multicolumn{1}{r}{\textbf{HSR-GloVe}} & \multicolumn{1}{r}{\textbf{GN-GloVe}} \\
           \toprule
   \textbf{Bias-by-projection} & 0.0375 & \underline{\textbf{0.0007}} & 0.0366 & 0.0218 & 0.0555 \\
    \textbf{SemBias} & 0.8023 & 0.8250 & 0.8432 & \textbf{0.8591} & \underline{0.9773} \\
    \textbf{SemBias (subset)} & 0.5750 & 0.3250 & \textbf{0.6500} & 0.1000 & \underline{0.7500} \\
      \bottomrule
    \end{tabular}%

    }
  \label{tab:gender_direction}%
\end{table}%

\paragraph{Bias-by-projection}

Bias-by-projection is the dot product between the target word and the gender direction $\overrightarrow{he}-\overrightarrow{she}$. Following \citet{gonen2019lipstick}, we report the average absolute bias-by-projection of the embedding of the top 500 male-biased words and the top 500 female-biased words. These words are determined according to the bias on the original GloVe embedding.

The first row in Table \ref{tab:gender_direction} shows the bias-by-projection, where the result marked in bold indicates the best result among all post-processing methods, and the result underlined is the globally best result. We could see that, while Hard-GloVe has the least bias-by-projection among all methods, our method is the runner-up and outperforms the word-vector-learning method GN-GloVe.

\paragraph{SemBias}

SemBias is an analogy task created by \citet{zhao2018learning} after SemEval 2012 Task2 \citep{jurgens2012semeval}. The task aims at identifying the gender-definition word pair from four pairs of words, including a gender-definition word pair, a gender-biased word pair, and two other pairs of words. The dataset contains 440 instances, in which 40 instances are not used during the training process of GN-GloVe, GP-GloVe, and HSR-GloVe. Hence the 40 instances (denoted by SemBias (subset)) are used to test the generalizability of the methods. For each instance, we calculate the subtraction $\overrightarrow{a}-\overrightarrow{b}$ for each word pair $(a,b)$, and the word pair with the highest cosine similarity between the subtraction and the gender direction $\overrightarrow{he}-\overrightarrow{she}$ is selected as the prediction based on the corresponding embedding.

The accuracy of identifying the gender-definition word pairs is reported in Table \ref{tab:gender_direction}. It is observed that our method is the best-performing post-processing method on SemBias.

\subsection{Gender-Biased Word Relation}

Gender-biased word relation tasks examine whether gender bias in word vector relation exists (see Section 2.1 for more details). We use the five gender-biased word relation tasks proposed by \citet{gonen2019lipstick}.

\begin{table}[ht]
  \centering
  \caption{Result of gender-biased word relation tasks}
\scalebox{0.60}{%
    \begin{tabular}{lrrrrr}
     \toprule
          &       & \multicolumn{3}{c}{\textbf{Post-Processing}} & \multicolumn{1}{c}{\textbf{WV-Learning}} \\
           \cmidrule(r){2-5} \cmidrule(r){6-6}
          & \multicolumn{1}{r}{\textbf{GloVe}} & \multicolumn{1}{r}{\textbf{Hard-GloVe}} & \multicolumn{1}{r}{\textbf{GP-GloVe}} & \multicolumn{1}{r}{\textbf{HSR-GloVe}} & \multicolumn{1}{r}{\textbf{GN-GloVe}} \\
               \toprule
  \textbf{GBWR-Clustering} & 1.0000 & \underline{\textbf{0.8050}} & 1.0000 & 0.9410 & 0.8560 \\
    \textbf{GBWR-Correlation} & 0.7727 & 0.6884 & 0.7700 & \underline{\textbf{0.6422}} & 0.7336 \\
    \textbf{GBWR-Profession} & 0.8200 & 0.7161 & 0.8102 & \underline{\textbf{0.6804}} & 0.7925 \\
    \textbf{GBWR-Association} & 2     & \underline{\textbf{1}} & 3     & \underline{\textbf{1}} & 3 \\
    \textbf{GBWR-Classification} & 0.9980 & 0.9068 & 0.9978 & \underline{\textbf{0.9055}} & 0.9815 \\

         \bottomrule
    \end{tabular}%

    }
  \label{tab:gender_word}%
\end{table}%

\paragraph{Clustering male- and female-biased words (GBWR-Clustering)}

This task takes the top 500 male-biased words and the top 500 female-biased words according to the original embedding and clusters them into two clusters using k-means \citep{lloyd1982least}. The purity \citep{Manning2008} against the original biased male and female clusters is shown in Table \ref{tab:gender_word}. We could see that HSR-GloVe is the runner-up among all post-processing methods.

\paragraph{Correlation between bias-by-projection and bias-by-neighbors (GBWR-Correlation)}

While we demonstrate the bias-by-projection in Section 4.1, bias-by-neighbors refers to the percentage of male/female socially-biased words among the k-nearest neighbors of the target word, where $k=100$. The GBWR-Correlation task calculates the Pearson correlation between the two biases.

In Table \ref{tab:gender_word}, it is observed that our proposed HSR-GloVe has the lowest Pearson correlation coefficient and outperforms all post-processing and word-vector-learning methods. Specifically, HSR brings about 6.72\% reduction of the correlation coefficient compared to the runner-up method, and it achieves 16.89\% reduction compared to the original GloVe embedding.

\begin{figure*}[htpb]

\centering

\begin{subfigure}[t]{.3\textwidth}
\centering
\includegraphics[width=\textwidth]{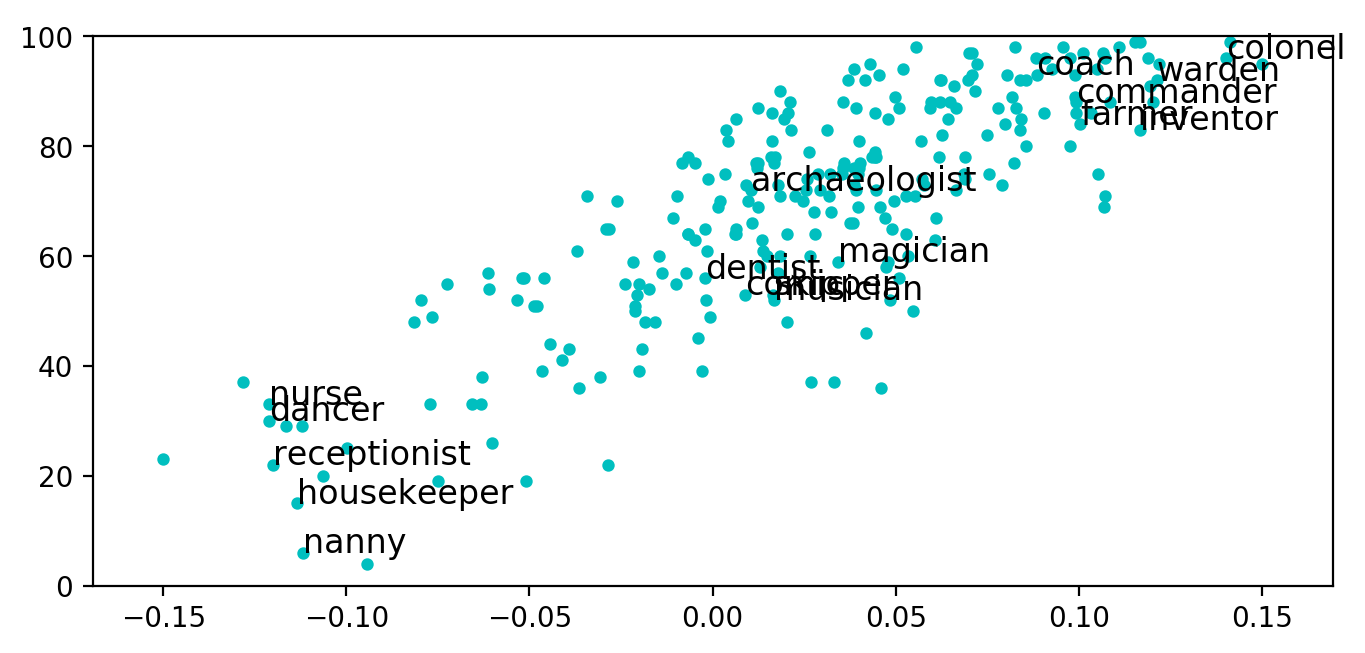}
\caption{Original GloVe}
\end{subfigure}
\quad
\begin{subfigure}[t]{.3\textwidth}
\centering
\includegraphics[width=\textwidth]{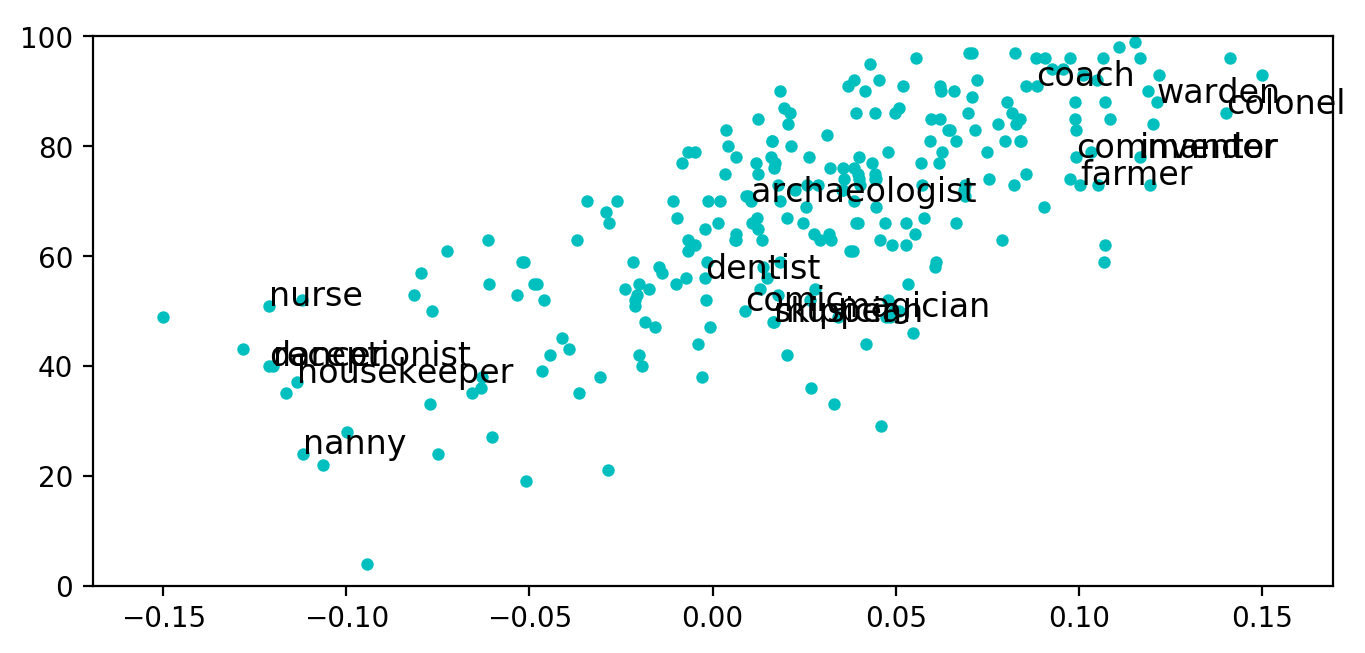}
\caption{Hard-GloVe}
\end{subfigure}
\quad
\begin{subfigure}[t]{.3\textwidth}
\centering
\includegraphics[width=\textwidth]{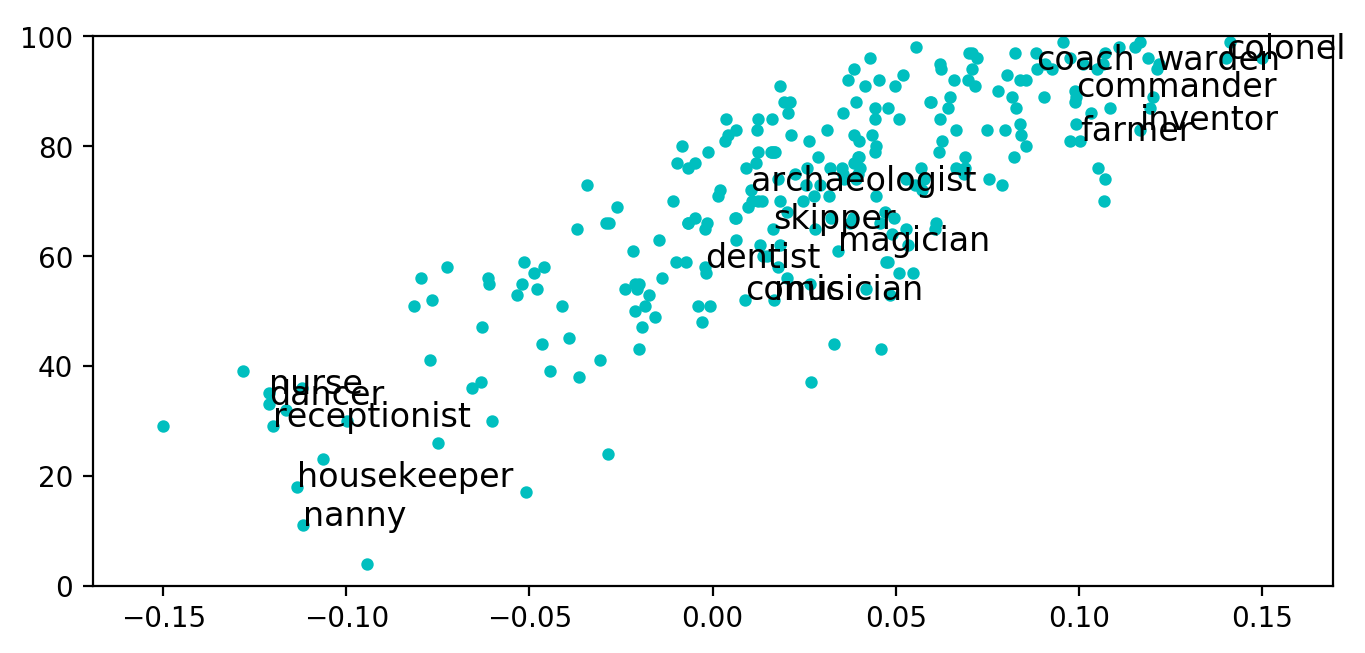}
\caption{GP-GloVe}
\end{subfigure}
\quad
\begin{subfigure}[t]{.3\textwidth}
\centering
\includegraphics[width=\textwidth]{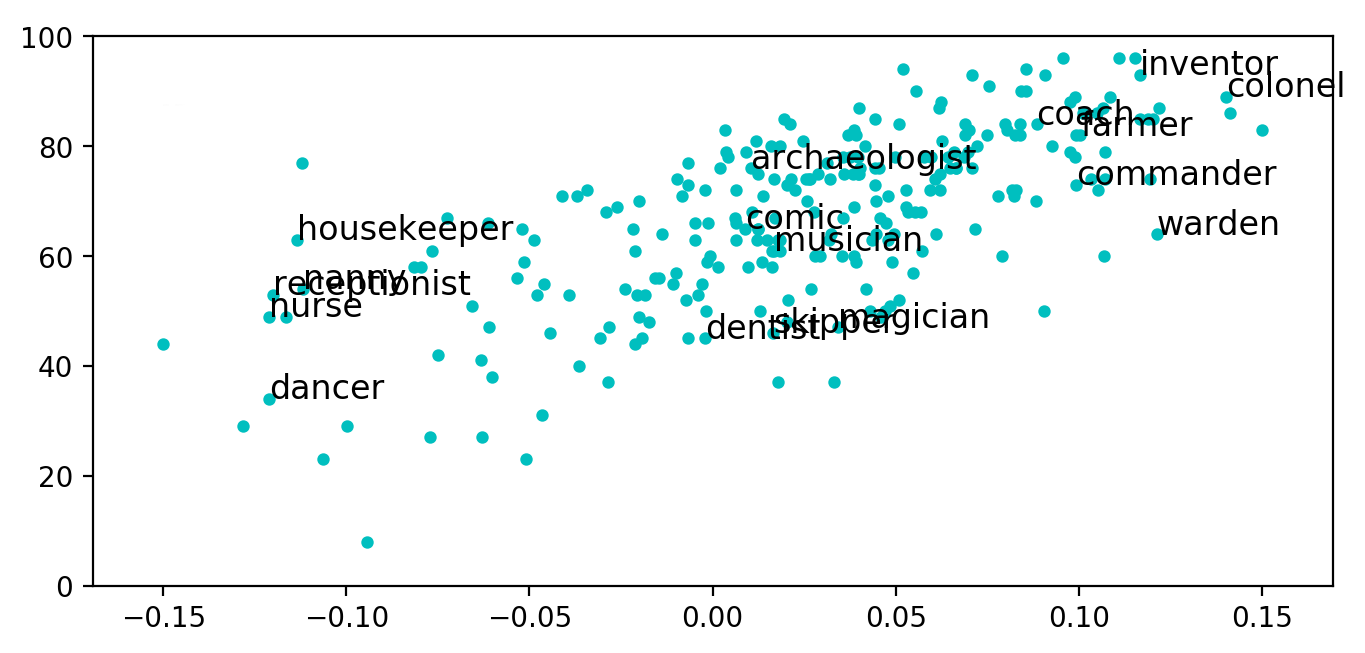}
\caption{HSR-GloVe}
\end{subfigure}
\quad
\begin{subfigure}[t]{.3\textwidth}
\centering
\includegraphics[width=\textwidth]{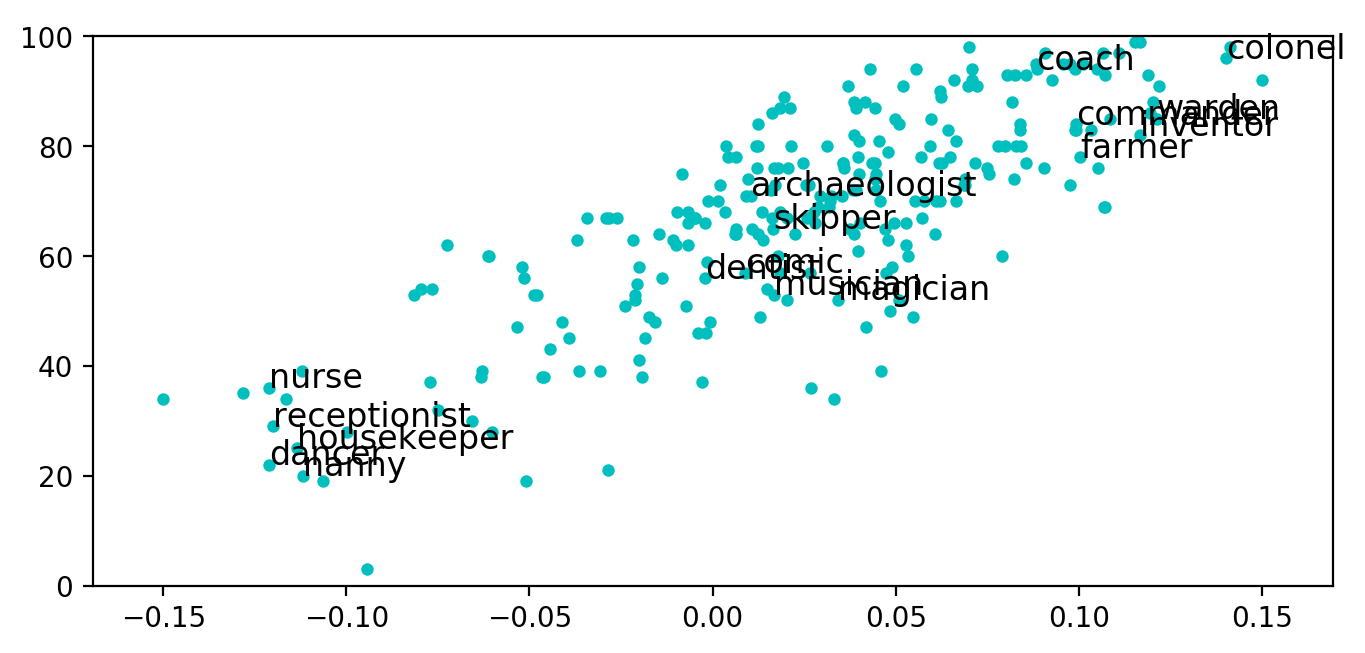}
\caption{GN-GloVe}
\end{subfigure}

\caption{The number of male neighbors for each profession against its original gender bias. Only a limited number of professions are shown on the plots to ensure readability.}
\label{fig:p}
\end{figure*}

\paragraph{Bias-by-neighbors of profession words (GBWR-Profession)}

For this task, we first calculate the 100 nearest neighbors of profession words adopted by \citet{bolukbasi2016man} and \citet{zhao2018learning}. Then, we compute the Pearson correlation between the number of male neighbors and the original bias.

The Pearson correlation coefficient is shown in Table \ref{tab:gender_word}. Again, our proposed HSR-GloVe has the lowest Pearson correlation coefficient and outperforms all other methods. Moreover, our method further reduces the correlation coefficient by 4.98\% compared to the runner-up method, and it results in 17.02\% reduction compared to the original GloVe embedding.

We also plot the number of male neighbors for each profession against its original gender bias in Figure \ref{fig:p}. We could see that for HSR-GloVe, the percentage of male neighbors is much closer to 50\% for words with different original gender bias. For example, after debiasing via HSR, the number of male neighbors of the female-biased word vector $\overrightarrow{nanny}$ and the male-biased word vector $\overrightarrow{warden}$ are both around 60, while in the original GloVe embedding, $\overrightarrow{nanny}$ has less than 10 male neighbors and $\overrightarrow{warden}$ has around 90 male neighbors.

\paragraph{Association between female/male and female/male-stereotyped words (GBWR-Association)} 

Following the Word Embedding Association Test (WEAT) created by \citet{caliskan2017semantics}, \citet{gonen2019lipstick} develop three association tasks which measure the association between female/male names and three groups of words including family and career words, arts and mathematics words, and arts and science words.

Table \ref{tab:gender_word} documents the number of significant p-values in the three tasks, and the p-values of the three tasks for each algorithm are reported in the appendix. HSR-GloVe and Hard-GloVe have the best performance that only one of the three p-values is significant.

\paragraph{Classifying previously female- and male-biased words (GBWR-Classification)}

In this task, the top 5000 female- and male-biased words (2500 for each gender) are sampled, and an SVM classifier is trained on a subset of 1000 words to predict the gender of the rest 4000 words.

The classification accuracy is reported in Table \ref{tab:gender_word}, we could see that HSR-GloVe has the least accuracy among all debiasing methods, indicating that it preserves the least gender bias.

\subsection{Lexical- and Sentence-Level Evaluation}

Besides the gender-debiasing quality, the general quality of word embedding is also crucial as it influences the performance of word embedding in many downstream NLP tasks. To test the gender-debiasing methods, we employ one lexical-level task: (a) word similarity task; and one sentence-level task (b) semantic textual similarity task.

\paragraph{Word similarity}

Word similarity tasks compare the similarity between embeddings of word pairs with human-annotated similarity scores. We use eight word similarity tasks to evaluate the proposed gender-debiasing method. The eight tasks include RG65 \citep{Rubenstein1965}, WordSim-353 \citep{Finkelstein2002}, Rare-words \citep{Luong2013}, MEN \citep{Bruni2014}, MTurk-287 \citep{Radinsky2011}, MTurk-771 \citep{halawi2012large}, SimLex-999 \citep{Hill2015}, and SimVerb-3500 \citep{Gerz2016}. For each task, the Spearman's rank correlation coefficient \citep{Myers1995} of the estimated rankings against the human rankings is calculated and shown in Table \ref{tab:word_sim}, where the result marked in bold is the best result among all post-processing methods, and the result underlined is the globally best result.

From the table, we could observe that HSR-GloVe has a dominantly better performance against all other post-processing methods as it obtains the best result in seven out of the eight tasks. Compared to the word-vector-learning method, HSR-GloVe still has the best performance by having five globally best coefficients out of the eight tasks. Overall, HSR-GloVe improves the coefficient of the eight word similarity tasks by 6.96\% on average compared to the prior best post-processing methods, and it results in 7.31\% improvement compared to the original embedding.

\begin{table}[htbp]
  \centering
  \caption{Spearman's rank correlation coefficient of word similarity tasks}
\scalebox{0.64}{%
    \begin{tabular}{lrrrrr}
    \toprule
          &       & \multicolumn{3}{c}{\textbf{Post-Processing}} & \multicolumn{1}{c}{\textbf{WV-Learning}} \\
          \cmidrule(r){2-5} \cmidrule(r){6-6}
          & \multicolumn{1}{r}{\textbf{GloVe}} & \multicolumn{1}{r}{\textbf{Hard-GloVe}} & \multicolumn{1}{r}{\textbf{GP-GloVe}} & \multicolumn{1}{r}{\textbf{HSR-GloVe}} & \multicolumn{1}{r}{\textbf{GN-GloVe}} \\
            \toprule
   \textbf{RG65} & 0.7540 & 0.7648 & 0.7546 & \underline{\textbf{0.7764}} & 0.7457 \\
    \textbf{WordSim-353} & 0.6199 & 0.6207 & 0.6003 & \underline{\textbf{0.6554}} & 0.6286 \\
    \textbf{RW} & 0.3722 & 0.3720 & 0.3450 & \textbf{0.3868} & \underline{0.3989} \\
    \textbf{MEN} & 0.7216 & 0.7212 & 0.6974 & \textbf{0.7353} & \underline{0.7446} \\
    \textbf{MTurk-287} & \textbf{0.6480} & 0.6468 & 0.6418 & 0.6335 & \underline{0.6617} \\
    \textbf{MTurk-771} & 0.6486 & 0.6504 & 0.6391 & \underline{\textbf{0.6652}} & 0.6619 \\
    \textbf{SimLex-999} & 0.3474 & 0.3501 & 0.3389 & \underline{\textbf{0.3971}} & 0.3700 \\
    \textbf{SimVerb-3500} & 0.2038 & 0.2034 & 0.1877 & \underline{\textbf{0.2635}} & 0.2219 \\

      \bottomrule
    \end{tabular}%
    }
  \label{tab:word_sim}%
\end{table}%

\paragraph{Semantic textual similarity}

The semantic textual similarity (STS) tasks examine the sentence-level effectiveness of word embeddings by measuring the degree of semantic equivalence between two texts \citep{Agirre2012}. The STS tasks we employ include 20 tasks from 2012 SemEval Semantic Related task (SICK) and SemEval STS tasks from 2012 to 2015 \citep{Marco2014,Agirre2012,Agirre2013,Agirre2014,Agirre2015}. For each task, we average the embedding of words in the sentence as the sentence embedding, and record the Pearson correlation coefficient of the estimated rankings of sentence similarity against the human rankings. In Table \ref{tab:STS}, we report the average Pearson correlation coefficient for the STS tasks each year (detailed results are in the appendix).

From Table \ref{tab:STS}, we could see that HSR-GloVe outperforms all post-processing and word-vector-learning gender-debiasing methods by performing the best globally on four out of the five tasks. For the SICK task, HSR-GloVe performs the best among all post-processing methods. On average, HSR-GloVe achieves an improvement of 14.19\% over the 20 STS tasks with respect to the prior best post-processing methods, and it attains 14.71\% improvement with respect to the original embedding.

\begin{table}[htbp]
  \centering
  \caption{Average Pearson correlation coefficient of semantic textual similarity tasks each year}
\scalebox{0.69}{%
           \begin{tabular}{lrrrrr}
           \toprule
          &       & \multicolumn{3}{c}{\textbf{Post-Processing}} & \multicolumn{1}{c}{\textbf{WV-Learning}} \\
          \cmidrule(r){2-5} \cmidrule(r){6-6}
          & \multicolumn{1}{r}{\textbf{GloVe}} & \multicolumn{1}{r}{\textbf{Hard-GloVe}} & \multicolumn{1}{r}{\textbf{GP-GloVe}} & \multicolumn{1}{r}{\textbf{HSR-GloVe}} & \multicolumn{1}{r}{\textbf{GN-GloVe}} \\
              \toprule
    \textbf{STS-2012} & 48.92 & 48.03 & 45.34 & \underline{\textbf{51.27}} & 50.90 \\
    \textbf{STS-2013} & 46.90 & 46.92 & 43.16 & \underline{\textbf{52.46}} & 48.71 \\
    \textbf{STS-2014} & 50.99 & 50.41 & 46.67 & \underline{\textbf{60.07}} & 54.16 \\
    \textbf{STS-2015} & 51.26 & 50.02 & 47.22 & \underline{\textbf{61.36}} & 53.49 \\
    \textbf{SICK} & 62.11 & 61.23 & 59.02 & \textbf{62.56} & \underline{63.58} \\
        \bottomrule
    \end{tabular}%
    }
  \label{tab:STS}%
\end{table}%

\subsection{Downstream Task: Coreference Resolution}
Lastly, we examine whether HSR-debiased word vectors reduce gender bias in the downstream task coreference resolution. Coreference resolution tasks strive to discover all textual mentions that refer to the same entity in a given text. Recently, \citet{zhao2018gender} show that word embedding is one of the causes of gender-biased identification outcomes of neural coreference resolution systems. We test the word embeddings on two coreference resolution tasks: OntoNotes 5.0 \citep{weischedel2013ontonotes} and WinoBias \citep{zhao2018gender}. OntoNotes 5.0 is a benchmark dataset for coreference resolution with texts from various genres. WinoBias dataset evaluates the level of gender bias in coreference resolution outcomes. The dataset is split into pro-stereotype (PRO) and anti-stereotype (ANTI) datasets. The PRO dataset contains sentences in which a gender pronoun (e.g. $she$) refers to a profession (e.g. $nurse$) that is biased toward the gender represented by the pronoun. The ANTI dataset is composed of the same sentences as the PRO dataset, except that the gender pronouns are reversed manually. A coreference resolution model passes the WinoBias test when the pro-stereotyped and anti-stereotyped coreference outcomes are made with the same accuracy \citep{zhao2018gender}.

\begin{table}[htbp]
  \centering
  \caption{Average F1 score of coreference resolution tasks}
\scalebox{0.64}{%
    \begin{tabular}{lrrrrr}
    \toprule
          &       & \multicolumn{3}{c}{\textbf{Post-Processing}} & \multicolumn{1}{c}{\textbf{WV-Learning}} \\
           \cmidrule(r){2-5} \cmidrule(r){6-6}
          & \multicolumn{1}{r}{\textbf{GloVe}} & \multicolumn{1}{r}{\textbf{Hard-GloVe}} & \multicolumn{1}{r}{\textbf{GP-GloVe}} & \multicolumn{1}{r}{\textbf{HSR-GloVe}} & \multicolumn{1}{r}{\textbf{GN-GloVe}} \\
          \toprule
    \textbf{OntoNotes-test} & \underline{\textbf{66.5}}  & 66.2  & 66.2  & 66.2  & 66.2 \\
    \textbf{WinoBias-PRO} & 76.2  & 70.6  & 71.2  & 63.0  & 72.4 \\
    \textbf{WinoBias-ANTI} & 46.0  & 54.9  & 52.4  & 61.0  & 51.9 \\
    \textbf{WinoBias-Avg} & 61.1  & 62.8  & 61.8  & 62.0  & 62.2 \\
    \textbf{WinoBias-Diff} & 30.2  & 15.7  & 18.8  & \underline{\textbf{2.0}}   & 20.5 \\
    \bottomrule
    \end{tabular}%
    }
  \label{tab:coref}%
\end{table}%

Following the experimental setting of \citet{zhao2018learning}, we train the end-to-end coreference resolution model \citep{lee2017end} using OntoNotes 5.0 train set for each of the embeddings, and we report the average F1 score on OntoNotes 5.0 test set, WinoBias PRO test set (Type 1), and WinoBias ANTI test set (Type 1). Furthermore, we also report the average (WinoBias-Avg) and the difference (WinoBias-Diff) between WinoBias PRO and ANTI test sets.

The experimental results of coreference resolution tasks are shown in Table \ref{tab:coref}. The results of GloVe, Hard-GloVe, and GN-GloVe are collected from \citet{zhao2018learning}. From the table, we could see that HSR-GloVe performs the best on the WinoBias task by achieving the lowest difference (WinoBias-Diff) between PRO and ANTI datasets. Specifically, HSR-GloVe reduces WinoBias-Diff by 87.26\% compared to the runner-up method Hard-GloVe and by 93.38\% compared to the original GloVe embedding. Furthermore, HSR-GloVe has an on-par performance on the OntoNotes 5.0 benchmark dataset compared to other baselines.

\section{Conclusion}

In this paper, we introduce a causal and simple gender-debiasing method which reduces gender bias during the post-processing period of word embeddings. Our proposed Half-Sibling Regression algorithm learns the spurious gender information via Ridge Regression and subtracts the learned gender information from the original non-gender-definition word vectors to obtain the gender-debiased word vectors. The experimental results on gender direction relation tasks, gender-biased word relation tasks, lexical- and sentence-level evaluation tasks, and coreference resolution tasks show that our proposed method (1) reduces gender bias in word vector relation as well as gender bias associated with gender direction, (2) enhances the lexical- and sentence-level quality of the word vectors, (3) diminishes gender bias in downstream tasks such as coreference resolution, and (4) consistently improves performance over the existing post-processing and word-vector-learning methods.

In the future, we will incorporate other supervised learning algorithms into the Half-Sibling framework and compare their performance to the Ridge Regression method proposed in this paper. Moreover, since our proposed algorithm is only suitable for alleviating gender bias of non-contextualized word embeddings, we will try to extend our method to debias contextualized word embeddings. We believe that HSR can be adopted as a post-processing method for contextualized word embeddings, and it is also possible to incorporate HSR during the training process of contextualized models. Last but not least, our method can be generalized to identify and mitigate other social biases such as racism in word vectors.

\paragraph{Acknowledgement} This work was supported by the Hong Kong Research Grants Council (Grant Numbers: 11500216, 11501414, 11507218). We appreciate the constructive comments from the anonymous reviewers. We thank Tianlin Liu for his helpful discussion and invaluable advice. We thank all the people who helped Zekun Yang flee from Hong Kong to Shenzhen on Nov. 12th, 2019 such that she could safely finish writing the camera-ready version of this paper.

\bibliography{3321.YangZ.bib}

\begin{thebibliography}{}

\bibitem[\protect\citeauthoryear{Agirre \bgroup et al\mbox.\egroup
  }{2012}]{Agirre2012}
Agirre, E.; Diab, M.; Cer, D.; and Gonzalez-Agirre, A.
\newblock 2012.
\newblock Semeval-2012 task 6: A pilot on semantic textual similarity.
\newblock In {\em Proceedings of the First Joint Conference on Lexical and
  Computational Semantics}, SemEval '12,  385--393.
\newblock Stroudsburg, PA, USA: Association for Computational Linguistics.

\bibitem[\protect\citeauthoryear{Agirre \bgroup et al\mbox.\egroup
  }{2013}]{Agirre2013}
Agirre, E.; Cer, D.; Diab, M.; Gonzalez-Agirre, A.; and Guo, W.
\newblock 2013.
\newblock Sem 2013 shared task: Semantic textual similarity.
\newblock In {\em Second Joint Conference on Lexical and Computational
  Semantics}, volume~1,  32--43.

\bibitem[\protect\citeauthoryear{Agirre \bgroup et al\mbox.\egroup
  }{2014}]{Agirre2014}
Agirre, E.; Banea, C.; Cardie, C.; Cer, D.; Diab, M.; Gonzalez-Agirre, A.; Guo,
  W.; Mihalcea, R.; Rigau, G.; and Wiebe, J.
\newblock 2014.
\newblock Semeval-2014 task 10: Multilingual semantic textual similarity.
\newblock In {\em Proceedings of the 8th international workshop on semantic
  evaluation},  81--91.

\bibitem[\protect\citeauthoryear{Agirre \bgroup et al\mbox.\egroup
  }{2015}]{Agirre2015}
Agirre, E.; Banea, C.; Cardie, C.; Cer, D.; Diab, M.; Gonzalez-Agirre, A.; Guo,
  W.; Lopez-Gazpio, I.; Maritxalar, M.; Mihalcea, R.; Rigaua, G.; Uriaa, L.;
  and Wiebeg, J.
\newblock 2015.
\newblock Semeval-2015 task 2: Semantic textual similarity, {E}nglish,
  {S}panish and pilot on interpretability.
\newblock In {\em Proceedings of the 9th international workshop on semantic
  evaluation},  252--263.

\bibitem[\protect\citeauthoryear{Bolukbasi \bgroup et al\mbox.\egroup
  }{2016}]{bolukbasi2016man}
Bolukbasi, T.; Chang, K.-W.; Zou, J.~Y.; Saligrama, V.; and Kalai, A.~T.
\newblock 2016.
\newblock Man is to computer programmer as woman is to homemaker? debiasing
  word embeddings.
\newblock In {\em Advances in neural information processing systems},
  4349--4357.

\bibitem[\protect\citeauthoryear{Bruni, Tran, and Baroni}{2014}]{Bruni2014}
Bruni, E.; Tran, N.~K.; and Baroni, M.
\newblock 2014.
\newblock Multimodal distributional semantics.
\newblock {\em Journal of Artificial Intelligence Research} 49(1):1--47.

\bibitem[\protect\citeauthoryear{Caliskan, Bryson, and
  Narayanan}{2017}]{caliskan2017semantics}
Caliskan, A.; Bryson, J.~J.; and Narayanan, A.
\newblock 2017.
\newblock Semantics derived automatically from language corpora contain
  human-like biases.
\newblock {\em Science} 356(6334):183--186.

\bibitem[\protect\citeauthoryear{Finkelstein \bgroup et al\mbox.\egroup
  }{2002}]{Finkelstein2002}
Finkelstein, L.; Gabrilovich, E.; Matias, Y.; Rivlin, E.; Solan, Z.; Wolfman,
  G.; and Ruppin, E.
\newblock 2002.
\newblock Placing search in context: the concept revisited.
\newblock {\em ACM Transactions on Information Systems} 20(1):116--131.

\bibitem[\protect\citeauthoryear{Ganguly \bgroup et al\mbox.\egroup
  }{2015}]{ganguly2015word}
Ganguly, D.; Roy, D.; Mitra, M.; and Jones, G.~J.
\newblock 2015.
\newblock Word embedding based generalized language model for information
  retrieval.
\newblock In {\em Proceedings of the 38th international ACM SIGIR conference on
  research and development in information retrieval},  795--798.
\newblock ACM.

\bibitem[\protect\citeauthoryear{Gerz \bgroup et al\mbox.\egroup
  }{2016}]{Gerz2016}
Gerz, D.; Vulic, I.; Hill, F.; Reichart, R.; and Korhonen, A.
\newblock 2016.
\newblock {S}im{V}erb-3500: a large-scale evaluation set of verb similarity.
\newblock In {\em Proceedings of the {EMNLP} 2016},  2173--2182.

\bibitem[\protect\citeauthoryear{Gonen and Goldberg}{2019}]{gonen2019lipstick}
Gonen, H., and Goldberg, Y.
\newblock 2019.
\newblock Lipstick on a pig: Debiasing methods cover up systematic gender
  biases in word embeddings but do not remove them.
\newblock In {\em Proceedings of the 2019 Conference of the {NAACL}: Human
  Language Technologies, Volume 1 (Long and Short Papers)},  609--614.

\bibitem[\protect\citeauthoryear{Halawi \bgroup et al\mbox.\egroup
  }{2012}]{halawi2012large}
Halawi, G.; Dror, G.; Gabrilovich, E.; and Koren, Y.
\newblock 2012.
\newblock Large-scale learning of word relatedness with constraints.
\newblock In {\em Proceedings of the 18th ACM SIGKDD international conference
  on Knowledge discovery and data mining},  1406--1414.
\newblock ACM.

\bibitem[\protect\citeauthoryear{Hill, Reichart, and Korhonen}{2015}]{Hill2015}
Hill, F.; Reichart, R.; and Korhonen, A.
\newblock 2015.
\newblock Simlex-999: Evaluating semantic models with (genuine) similarity
  estimation.
\newblock {\em Computational Linguistics} 41(4):665--695.

\bibitem[\protect\citeauthoryear{Hoerl and Kennard}{1970}]{hoerl1970ridge}
Hoerl, A.~E., and Kennard, R.~W.
\newblock 1970.
\newblock Ridge regression: Biased estimation for nonorthogonal problems.
\newblock {\em Technometrics} 12(1):55--67.

\bibitem[\protect\citeauthoryear{Jurgens \bgroup et al\mbox.\egroup
  }{2012}]{jurgens2012semeval}
Jurgens, D.~A.; Turney, P.~D.; Mohammad, S.~M.; and Holyoak, K.~J.
\newblock 2012.
\newblock Semeval-2012 task 2: Measuring degrees of relational similarity.
\newblock In {\em Proceedings of the First Joint Conference on Lexical and
  Computational Semantics-Volume 1: Proceedings of the main conference and the
  shared task, and Volume 2: Proceedings of the Sixth International Workshop on
  Semantic Evaluation},  356--364.
\newblock Association for Computational Linguistics.

\bibitem[\protect\citeauthoryear{Kaneko and Bollegala}{2019}]{Kaneko:ACL:2019}
Kaneko, M., and Bollegala, D.
\newblock 2019.
\newblock Gender-preserving debiasing for pre-trained word embeddings.
\newblock In {\em Proceedings of the 57th Annual Meeting of the Association for
  Computational Linguistics}.

\bibitem[\protect\citeauthoryear{Lee \bgroup et al\mbox.\egroup
  }{2017}]{lee2017end}
Lee, K.; He, L.; Lewis, M.; and Zettlemoyer, L.
\newblock 2017.
\newblock End-to-end neural coreference resolution.
\newblock In {\em Proceedings of the {EMNLP} 2017},  188--197.

\bibitem[\protect\citeauthoryear{Lloyd}{1982}]{lloyd1982least}
Lloyd, S.
\newblock 1982.
\newblock Least squares quantization in pcm.
\newblock {\em IEEE transactions on information theory} 28(2):129--137.

\bibitem[\protect\citeauthoryear{Luong, Socher, and Manning}{2013}]{Luong2013}
Luong, M.; Socher, R.; and Manning, C.~D.
\newblock 2013.
\newblock Better word representations with recursive neural networks for
  morphology.
\newblock In {\em Proceedings of the {CoNLL} 2013}.

\bibitem[\protect\citeauthoryear{Manning, Raghavan, and
  Sch\"utze}{2008}]{Manning2008}
Manning, C.~D.; Raghavan, P.; and Sch\"utze, H.
\newblock 2008.
\newblock {\em Introduction to Information Retrieval}.
\newblock Cambridge University Press.

\bibitem[\protect\citeauthoryear{Marelli \bgroup et al\mbox.\egroup
  }{2014}]{Marco2014}
Marelli, M.; Menini, S.; Baroni, M.; Bentivogli, L.; Bernardi, R.; and
  Zamparelli, R.
\newblock 2014.
\newblock A {SICK} cure for the evaluation of compositional distributional
  semantic models.
\newblock In {\em Proceedings of the Ninth International Conference on Language
  Resources and Evaluation}.
\newblock European Language Resources Association (ELRA).

\bibitem[\protect\citeauthoryear{Mikolov \bgroup et al\mbox.\egroup
  }{2013}]{Mikolov2013}
Mikolov, T.; Sutskever, I.; Chen, K.; Corrado, G.~S.; and Dean, J.
\newblock 2013.
\newblock Distributed representations of words and phrases and their
  compositionality.
\newblock In Burges, C. J.~C.; Bottou, L.; Welling, M.; Ghahramani, Z.; and
  Weinberger, K.~Q., eds., {\em Advances in Neural Information Processing
  Systems 26}. Curran Associates, Inc.
\newblock  3111--3119.

\bibitem[\protect\citeauthoryear{Myers and Well}{1995}]{Myers1995}
Myers, J.~L., and Well, A.~D.
\newblock 1995.
\newblock {\em {Research Design \& Statistical Analysis}}.
\newblock Routledge, 1 edition.

\bibitem[\protect\citeauthoryear{Park, Shin, and Fung}{2018}]{park2018reducing}
Park, J.~H.; Shin, J.; and Fung, P.
\newblock 2018.
\newblock Reducing gender bias in abusive language detection.
\newblock In {\em Proceedings of the {EMNLP} 2018},  2799--2804.

\bibitem[\protect\citeauthoryear{Pennington, Socher, and
  Manning}{2014}]{Pennington2014}
Pennington, J.; Socher, R.; and Manning, C.~D.
\newblock 2014.
\newblock Glove: Global vectors for word representation.
\newblock In {\em Proceedings of the {EMNLP} 2014},  1532--1543.

\bibitem[\protect\citeauthoryear{Radinsky \bgroup et al\mbox.\egroup
  }{2011}]{Radinsky2011}
Radinsky, K.; Agichtein, E.; Gabrilovich, E.; and Markovitch, S.
\newblock 2011.
\newblock A word at a time: Computing word relatedness using temporal semantic
  analysis.
\newblock In {\em Proceedings of the 20th International World Wide Web
  Conference},  337--346.

\bibitem[\protect\citeauthoryear{Rubenstein and
  Goodenough}{1965}]{Rubenstein1965}
Rubenstein, H., and Goodenough, J.~B.
\newblock 1965.
\newblock Contextual correlates of synonymy.
\newblock {\em Communications of the ACM} 8(10):627--633.

\bibitem[\protect\citeauthoryear{Sch{\"o}lkopf \bgroup et al\mbox.\egroup
  }{2016}]{Scholkopf2016}
Sch{\"o}lkopf, B.; Hogg, D.~W.; Wang, D.; Foreman-Mackey, D.; Janzing, D.;
  Simon-Gabriel, C.; and Peters, J.
\newblock 2016.
\newblock Modeling confounding by half-sibling regression.
\newblock {\em Proceedings of the National Academy of Sciences}
  113(27):7391--7398.

\bibitem[\protect\citeauthoryear{Tang \bgroup et al\mbox.\egroup
  }{2014}]{tang2014learning}
Tang, D.; Wei, F.; Yang, N.; Zhou, M.; Liu, T.; and Qin, B.
\newblock 2014.
\newblock Learning sentiment-specific word embedding for twitter sentiment
  classification.
\newblock In {\em Proceedings of the 52nd Annual Meeting of the Association for
  Computational Linguistics (Volume 1: Long Papers)},  1555--1565.

\bibitem[\protect\citeauthoryear{Weischedel \bgroup et al\mbox.\egroup
  }{2013}]{weischedel2013ontonotes}
Weischedel, R.; Palmer, M.; Marcus, M.; Hovy, E.; Pradhan, S.; Ramshaw, L.;
  Xue, N.; Taylor, A.; Kaufman, J.; Franchini, M.; et~al.
\newblock 2013.
\newblock Ontonotes release 5.0 ldc2013t19.
\newblock {\em Linguistic Data Consortium, Philadelphia, PA} 23.

\bibitem[\protect\citeauthoryear{Yang \bgroup et al\mbox.\egroup
  }{2016}]{yang2016stacked}
Yang, Z.; He, X.; Gao, J.; Deng, L.; and Smola, A.
\newblock 2016.
\newblock Stacked attention networks for image question answering.
\newblock In {\em Proceedings of the IEEE conference on computer vision and
  pattern recognition},  21--29.

\bibitem[\protect\citeauthoryear{Zhao \bgroup et al\mbox.\egroup
  }{2018a}]{zhao2018gender}
Zhao, J.; Wang, T.; Yatskar, M.; Ordonez, V.; and Chang, K.-W.
\newblock 2018a.
\newblock Gender bias in coreference resolution: Evaluation and debiasing
  methods.
\newblock In {\em Proceedings of the 2018 Conference of the {NAACL}: Human
  Language Technologies, Volume 2 (Short Papers)},  15--20.

\bibitem[\protect\citeauthoryear{Zhao \bgroup et al\mbox.\egroup
  }{2018b}]{zhao2018learning}
Zhao, J.; Zhou, Y.; Li, Z.; Wang, W.; and Chang, K.-W.
\newblock 2018b.
\newblock Learning gender-neutral word embeddings.
\newblock In {\em Proceedings of the {EMNLP} 2018},  4847--4853.

\end{thebibliography}
\bibliographystyle{aaai}

\includepdf[pages={1}]{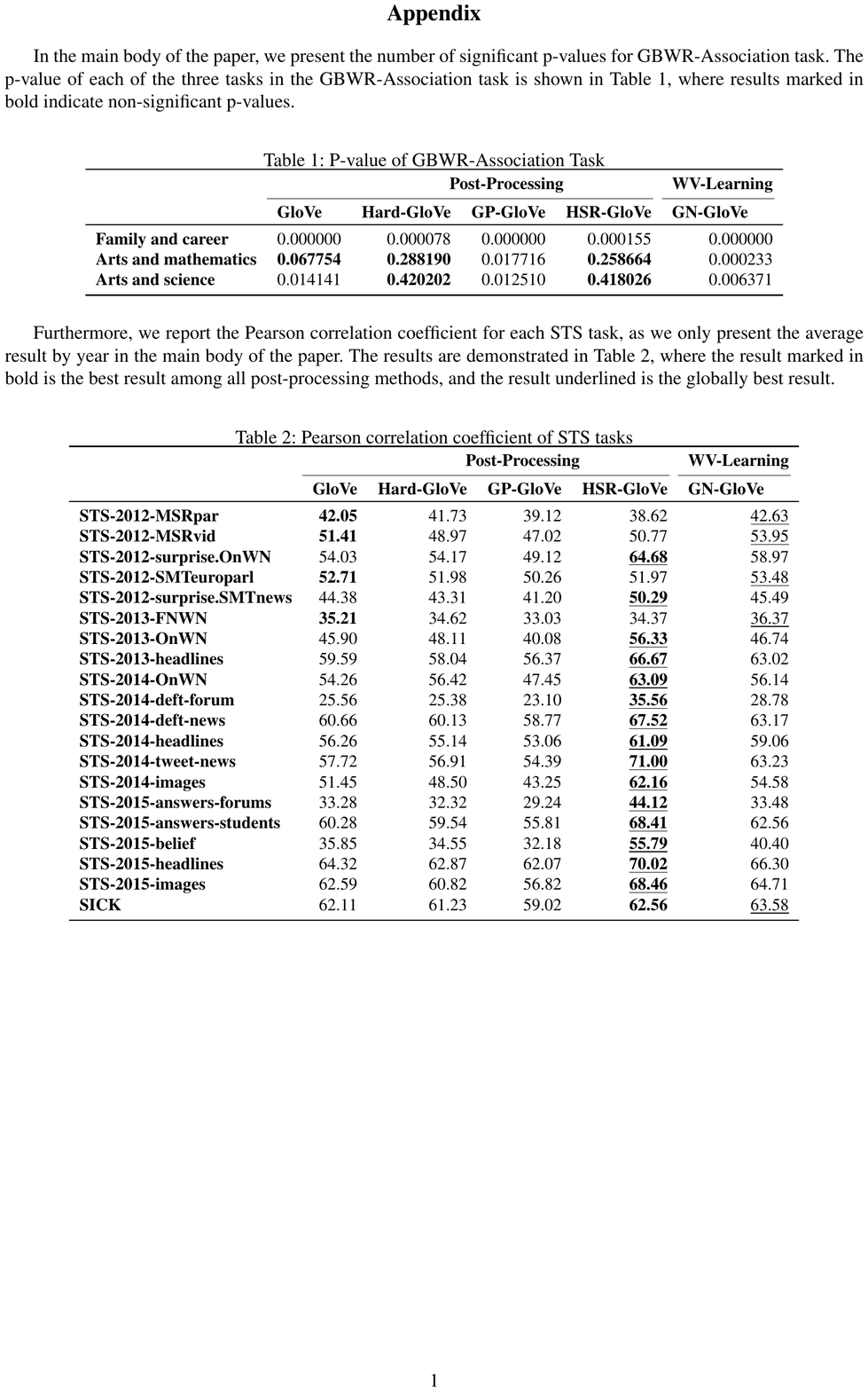}

\end{document}